\documentclass[letter,10pt,conference]{ieeeconf}
\IEEEoverridecommandlockouts
\title{
Motion planning for hundreds of floating robots}
\author{Anonymous Authors}

\usepackage{amsmath,amssymb,mathtools,amsthm}
\usepackage{mathtools}
\usepackage{physics}
\usepackage{empheq}
\usepackage{enumerate}
\let\labelindent\relax
\usepackage{enumitem}
\usepackage{pifont}
\usepackage[hidelinks]{hyperref} 
\usepackage[capitalize]{cleveref}
\usepackage{pgfplots}
\usepackage{pgfplotstable}
\usepgfplotslibrary{groupplots}
\usepgfplotslibrary{fillbetween}
\pgfplotsset{compat=1.17}
\usepackage{multirow}
\usepackage{xcolor}
\usepackage{colortbl}
\usepackage{adjustbox}
\usepackage{xspace}
\usepackage[T1]{fontenc}     
\usepackage[utf8]{inputenc}
\usepackage{tcolorbox}
\usepackage{cite} 
\usepackage{capt-of}
\usepackage{cuted}

\usepackage{booktabs}

\usepackage{algorithm}
\usepackage{algpseudocode}
\usepackage[framemethod=tikz]{mdframed}
\usepackage{pgfplots}
\usepackage{tikzpeople}
\usepackage{amsmath}
\usepackage{pgfplots}
\pgfplotsset{compat=1.18} 

\usepackage{tcolorbox}
\crefname{myalgorithm}{Algorithm}{Algorithms}
\crefname{figure}{Fig.}{Figs.}
\Crefname{figure}{Fig.}{Figs.}
\Crefname{experiment}{Experiment}{Experiments}
\Crefname{appendix}{Appendix}{Appendices}

\definecolor{truegreen}{RGB}{0,150,0}
\definecolor{noisyblue}{RGB}{31,119,180}
\definecolor{filteredorange}{RGB}{255,127,14}
\pgfplotsset{
  myaxis/.style={
    width=\linewidth,
    height=5cm,
    grid=both,
    grid style={line width=.2pt, opacity=.4},
    tick align=outside,
    tick style={black},
    label style={font=\small},
    ticklabel style={font=\scriptsize},
    scaled x ticks=false,
    scaled y ticks=false,
    xlabel style={yshift=6pt},   % shift upward (closer to grid)
    ylabel style={xshift=6pt},   % shift right (closer to grid)
    yticklabel style={
      /pgf/number format/fixed,
      /pgf/number format/precision=2
    }
  }
}

\usepackage{makecell}

\theoremstyle{definition}
\newtheorem{experiment}{Experiment}
\newtheorem{remark}{Remark}
\usepackage[acronym]{glossaries}
\newacronym{scp}{SCP}{sequential convex programming}
\newacronym{qp}{QP}{quadratic programs}
\newcommand{\citep}[1]{\cite{#1}}
\newcommand{\citet}[1]{\cite{#1}}

\begin{document}

\author{
    Jan Kamm$^\ast$, Antonio Terpin$^\ast$, Raffaello D'Andrea, Aswin Ramachandran
    \thanks{$^\ast$: Equal contribution. All authors are with the Institute for Dynamic Systems and Control, ETH Z\"urich. Corresponding authors: Antonio Terpin (aterpin@ethz.ch) and Jan Kamm (jakamm@ethz.ch).}
    \thanks{Accepted to the 2026 IEEE/RSJ International Conference on Intelligent Robots and Systems (IROS 2026).}
}

\maketitle
\thispagestyle{empty}
\pagestyle{empty}

\begin{strip}
\vspace{-1.7 cm}
\centering
\begin{minipage}{\linewidth}
\includegraphics[width=\textwidth]{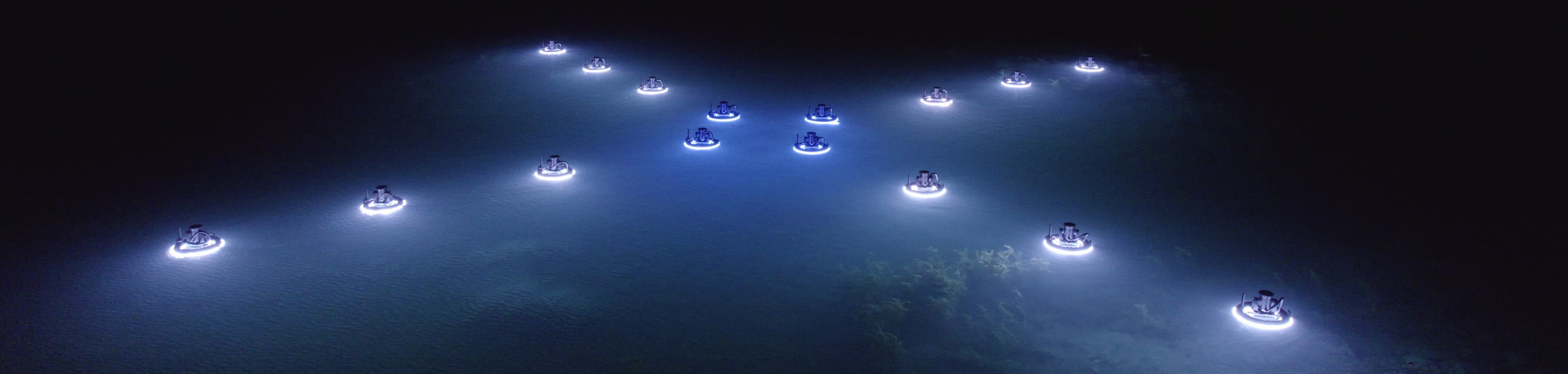}
\end{minipage}
  \captionof{figure}{Floating robots in formation on Lake Z\"urich. In this paper, we present a planning pipeline that generates dynamically feasible, collision-free trajectories for hundreds of robots fast enough for iterative show design.}
  \label{fig:planning-cover}
\end{strip}

\begin{abstract}
Planning collision-free motion for large robot fleets is difficult because collision avoidance induces strong inter-agent coupling that grows rapidly with team size. We consider omnidirectional floating robots on water, where choreographies are specified by sparse keyframes and an interactive tool must generate trajectories within seconds, even when transitions span minutes and thousands of time steps. We propose a scalable pipeline that builds a collision graph from an initialization, decomposes the coupled problem into interaction clusters, and solves clusters independently (and in parallel) with robustness mechanisms for common decomposition pathologies.
We validate the approach in simulations up to 500 robots. The synthesized trajectories have also been deployed in two real-world demonstrations, on Lake Z\"urich with a fleet of 24 \textit{Way of Water} crafts and at the Time Space Existence 2025 Venice Biennale.
\end{abstract}

\section{Introduction}
\label{sec:intro}

Planning collision-free motion for fleets of hundreds of robots remains challenging because collision avoidance introduces strong inter-agent coupling that grows rapidly (often quadratically) with fleet size. We study this problem in the context of floating robots operating on water (see \cref{fig:planning-cover}), where a show is specified by sparse \emph{keyframes}, identified by a specific fleet configuration. The planning problem is to compute, for every robot, a collision-free trajectory that is dynamically feasible (i.e., trackable by a low-level controller to $\mathrm{cm}$ accuracy) and that satisfies the boundary conditions at the beginning and end of the transition. 
For an interactive show-planning tool, a designer will iteratively adjust sparse keyframes and expect near-immediate feedback on the resulting fleet motion \citep{ramachandranvenkatapathy2026wayofwater}. Since consecutive keyframes can be separated by several minutes, a single transition may span thousands of planning steps. To render trajectories smoothly and to match the execution/preview rate, the planner must therefore produce dynamically feasible, collision-free trajectories over thousands of time steps within a few seconds per each transition between keyframes.
In contrast to aerial drone choreography, where robots move in a largely unobstructed \emph{three-dimensional} workspace and can use vertical maneuvers (e.g., temporary altitude offsets) to deconflict dense transitions, floating robots are constrained to the water surface and therefore evolve on a planar manifold. As a result, coordination and collision avoidance become intrinsically harder for surface fleets than for 3D aerial displays, where large multi-robot choreography has been demonstrated with integrated planning and collision-avoidance pipelines \citep{alonso-mora_multi-robot_2015,10.1145/2858036.2858353}.

\Gls*{scp} is a widely used workhorse for generating smooth, dynamically feasible, collision-avoiding trajectories by iteratively solving convex subproblems that form local inner approximations of nonconvex dynamics and collision constraints (often coupled with trust regions and/or penalty continuation). Conceptually, \gls*{scp} traces back to classical nonlinear-programming work on inner-approximation methods \citep{marks_technical_1978}. In robotics, early demonstrations include time-optimal car trajectory planning \citep{tran_application_2009} and coordinated multi-vehicle 3D-trajectory generation for fleets of flying drones \citep{augugliaro_generation_2012}. Subsequent work further systematized and popularized \gls*{scp} for motion planning among obstacles---notably through TrajOpt-style formulations that pair convexification with practical penalty continuation and efficient collision checking \citep{schulman_finding_2013,schulman_motion_2014}. In multi-robot and formation settings, related \gls*{scp} formulations have been used to compute coordinated 3D trajectories and to navigate robot teams among static and dynamic obstacles \citep{alonso-mora_multi-robot_2015}.
\begin{figure*}
\centering
\includegraphics[width=.9\linewidth]{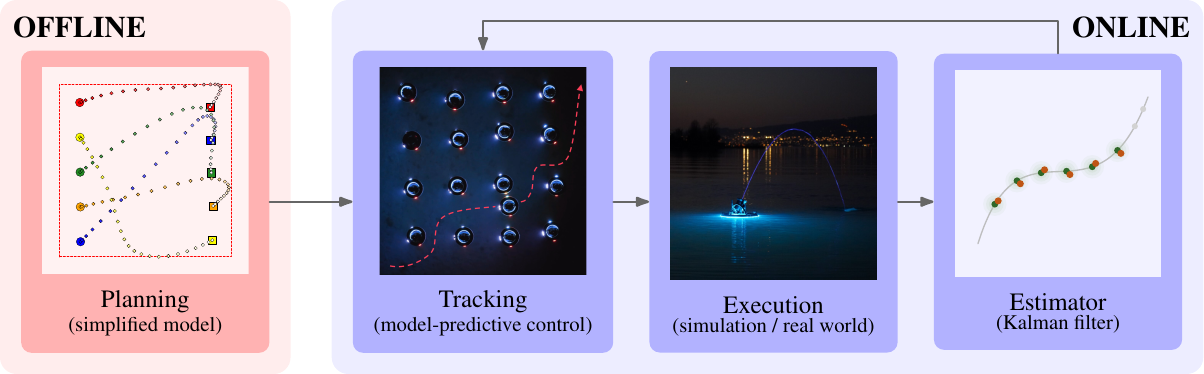}
\vspace{-.25cm}
    \caption{Schematic of the deployment procedure. An offline planner generates references using a simplified model, while an online model-predictive controller tracks them during execution with estimator feedback to handle model mismatch in simulation and in the real world.}
    \label{fig:offline-online}
\end{figure*}
Beyond \gls*{scp}, several complementary paradigms address multi-agent collision avoidance, each with its own tradeoffs. Reciprocal-velocity-obstacle methods such as ORCA \citep{siciliano_reciprocal_2011} provide distributed and reactive avoidance with formal pairwise safety guarantees under modeling assumptions, but no guarantee that robots reach the target states in the given time period. On discrete domains (e.g., graphs), conflict-based search tackles combinatorial coordination explicitly \citep{sharon_conflict-based_2015}. Within continuous trajectory optimization, planners such as CHOMP and STOMP handle obstacles primarily through smooth cost terms and stochastic or gradient-based updates rather than feasibility-focused convexification \citep{ratliff_chomp_2009,kalakrishnan_stomp_2011}, and other methods have approached the problem with distributed feedback optimization \citep{terpin2022distributed}. More recently, scalable interaction modeling has been pursued via learning-based decentralized policies \citep{chen_decentralized_2017} and neural control-barrier-function approaches \citep{zhang_neural_2023}. However, because continuous-time motion planning pipelines often require strict enforcement of boundary conditions and safety constraints (e.g., collision avoidance), \gls*{scp} remains a widely used workhorse.

A widely adopted optimization pipeline for multi-agent trajectory generation and execution is the one adopted by \citet{augugliaro_generation_2012}, which plans collision-free 3D-trajectories for fleets of flying drones using a simplified translational model, to be tracked by a low-level controller; see \cref{fig:offline-online}. However, this modeling simplification alone does not resolve the central scalability bottleneck: as the fleet size grows, collision avoidance induces rapidly increasing inter-agent coupling (and, in many formulations, a quadratic number of pairwise interactions), which can make \gls*{scp}-based planning computationally demanding at scale.
In fact, these \gls*{scp} formulations rely on the solution of several \gls*{qp} with complexity depending on the sparsity of the matrices involved~\citep{stellato_osqp_2020}. The number of collision constraints scales as~$\binom{N}{2}K$, with $N$ robots and $K$ discretized time steps. For $N=500, K = 1000$, the typical \gls*{qp}s in these settings have approximately hundreds of millions of non-zero entries, an intractable amount, in particular for surface robots that cannot perform 3D maneuvers.

\begin{mdframed}[hidealllines=true,backgroundcolor=blue!5]
\noindent\textbf{Contributions.}
We engineer and validate a scalable planning pipeline for large surface-robot fleets specified by sparse keyframes which is orders of magnitude faster than the traditional \gls*{scp} and more robust, producing collision-free trajectories within a few seconds on all the tested scenarios. Specifically,
\begin{itemize}[leftmargin=*]
    \item we introduce a hierarchical planner that builds a collision graph from an initialization and decomposes the problem into interaction clusters, enabling independent (and parallel) subproblem solves;
    \item we introduce robustness and efficiency mechanisms for the decomposition (fast graph/cluster updates and handling of common pathologies such as cyclic inter-cluster dependencies), with ablations isolating the effect of each design choice; and
    \item we improve cluster-level trajectory optimization introducing a reformulation in terms of the objective variable to improve the sparsity of the linearized \gls*{qp} and a dynamic time-scale selection to improve the robustness of the solver.
\end{itemize}
We validated the full system in simulations up to 500 robots and 1000 time steps, and the synthesized trajectories have been deployed in two real-world demonstrations, on Lake Z\"urich and at the Time Space Existence 2025 Venice Biennale, using the robotic platform Way of Water \citep{ramachandranvenkatapathy2026wayofwater}.
\end{mdframed}
\section{The planning problem}
\label{sec:allocation}
\begin{figure*}
    \centering
    \includegraphics[width=\textwidth]{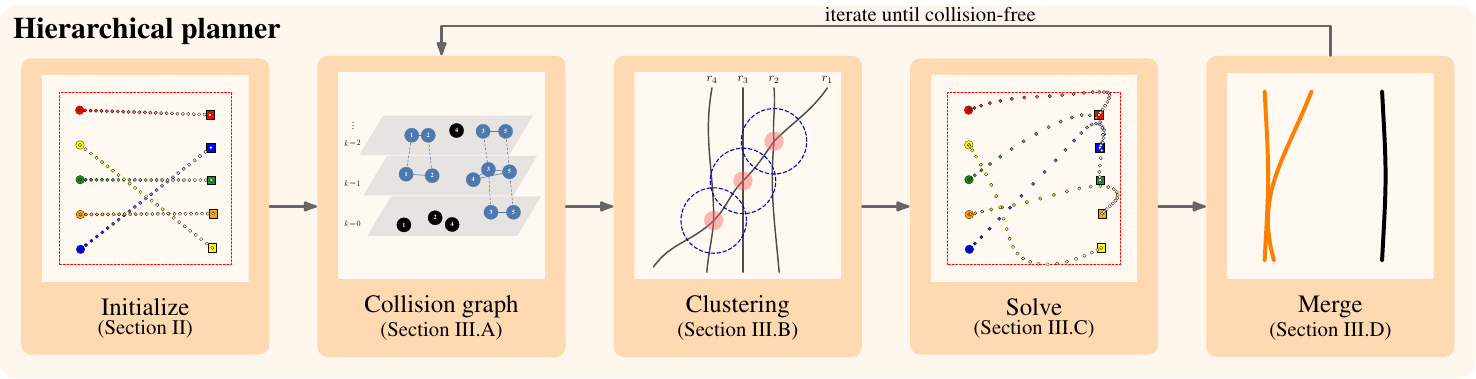}
    \vspace{-.5cm}
    \caption{Hierarchical solver pipeline overview.}
    \label{fig:pipeline}
\end{figure*}
\begin{figure}
    \centering
    \includegraphics[width=\linewidth]{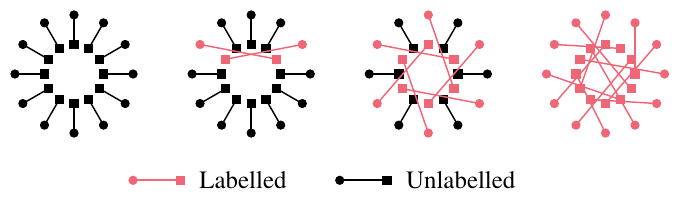}
    \vspace{-.75cm}
    \caption{Effect of the labelled fraction $|\mathcal{L}|/N$ on the allocation-induced straight-line initialization.
    As more robots are manually pinned, the remaining degrees of freedom shrink, and crossings become more frequent.}
    \label{fig:allocation-labelled-fraction}
\end{figure}
Our planning problem considers $N$ robots moving in $\mathbb{R}^2$ from initial positions
$\{p_0^{(i)}\}_{i=1}^N$ to a set of goal positions $\{p_1^{(i)}\}_{i=1}^N$ over a fixed horizon of $T$ seconds, 
while avoiding collisions. We discretize the horizon into $K$ steps of duration $h=T/K$ seconds and let $p^{(i)}[k]$ denote the position of robot $i$ at the discrete time index $k$.
Collision avoidance is enforced via the constraint
$\|p^{(i)}[k]-p^{(j)}[k]\|_2 \ge R$ for all $i\neq j$ and all $k$, where $R$ is a safety radius derived
from the robot geometry; in this work we set $R=0.8~\mathrm{m}$. In our setting, the floating robots are omnidirectional and rapidly reach the commanded speeds. Thus, we can plan directly in a single-integrator model without imposing higher-order smoothness (e.g., jerk) constraints. The goal of the planner is to find velocity controls $(v^{(i)}[k])_{k = 0, \ldots, K-1}^{i = 1, \ldots, N}$ so that the resulting fleet trajectory $\tau = (p^{(i)}[k])_{k = 0, \ldots, K}^{i = 1, \ldots, N}$ is collision free and $p^{(i)}[0] = p_0^{(i)}, p^{(i)}[K] = p_1^{(i)}$ for all $i = 1, \ldots, N$.

In a typical choreography, only a subset of the robots is \emph{manually} matched to specific target locations.
Let $\mathcal{L}\subseteq\{1,\ldots,N\}$ be the set of labelled robots, $\mathcal{U}=\{1,\ldots,N\}\setminus\mathcal{L}$
the unlabelled robots, and let the manual matching be encoded by an injective map
$\sigma:\mathcal{L}\to\{1,\ldots,N\}$. We compute a complete allocation by solving the
minimum-cost assignment problem (equivalently, an optimal transport problem between uniform discrete measures \cite{mirsky1963results,terpin_dynamic_2024})
\begin{equation}
\label{eq:allocation}
\begin{aligned}
\underset{\gamma_{ij}\ge 0}{\mathrm{min}}\quad \sum_{i=1}^N\sum_{j=1}^N c_{ij}\gamma_{ij}\quad
\mathrm{s.t.}\quad \sum_{i=1}^N \gamma_{ij}=\sum_{j=1}^N \gamma_{ij}=\frac{1}{N},
\end{aligned}
\end{equation}
with costs
\begin{equation}
\label{eq:cost}
c_{ij}=
\begin{cases}
0 & \text{if } i\in\mathcal{L}\ \text{and}\ j=\sigma(i),\\
+\infty & \text{if } i\in\mathcal{L}\ \text{and}\ j\neq\sigma(i),\\
\|p_0^{(i)}-p_1^{(j)}\|_2 & \text{if } i\in\mathcal{U}.
\end{cases}
\end{equation}
The $+\infty$ costs enforce the prescribed matches for labelled robots; for unlabelled robots, the Euclidean
cost yields assignments with small total displacement, which empirically reduces large swaps and simplifies
subsequent collision-free trajectory generation \cite{terpin_dynamic_2024,kuhn1955hungarian}; see \cref{fig:allocation-labelled-fraction}.

\begin{remark}
Increasing the fraction of labelled robots, i.e., enlarging $\mathcal{L}$, provides a simple knob that trades off
designer intent against automatic regularization. At $\mathcal{L} = \emptyset$, the
Euclidean cost in~\eqref{eq:cost} yields a small-displacement assignment
that avoids swaps and minimizes collisions. At $\mathcal{L} = \{1, \dots, N\}$, the assignment is fully user-prescribed, which can be arbitrarily challenging for solvers; robustness to such cases is an important feature of our pipeline. In practice, we observed that allocating as little as $10\%$ of the robots randomly, while allocating the remainder via \eqref{eq:allocation}, already injects enough entropy for visually appealing trajectories.
\end{remark}
For solving the linear program \eqref{eq:allocation} we use the Hungarian algorithm \cite{kuhn1955hungarian}, which results in sub-$\mathrm{ms}$ solve time for fleets of $500$ robots in the most challenging configuration ($0\%$ labelled robots).
Given the resulting allocation, we construct an initial (generally not collision-free) trajectory $\tau_0 = (p^{(i)}[k])^{i = 1, \ldots, N}_{k = 0, \ldots, K}$ by linear interpolation between initial and final positions.

% Hierarchical Solver --------------------------------
\section{Hierarchical Solver}
\label{sec:hierarchical}

At every timestep robots are spread in space and collisions have high spatio-temporal locality. To exploit this structure, we (i) identify \emph{collision clusters} (groups of interacting robots), (ii) convert each cluster into a smaller \emph{subproblem} that can be solved independently. Given the initialization $\tau_0$ from \cref{sec:allocation}, the hierarchical solver thus produces a sequence of fleet trajectories $\tau_1, \tau_2, \ldots$ until it obtains one that is collision free; see \cref{fig:pipeline}. In the remainder of this section, we describe the key components of the pipeline.

% -----------------------------------------------------------------------
\begin{figure}[t]
    \includegraphics[width=0.8\columnwidth]{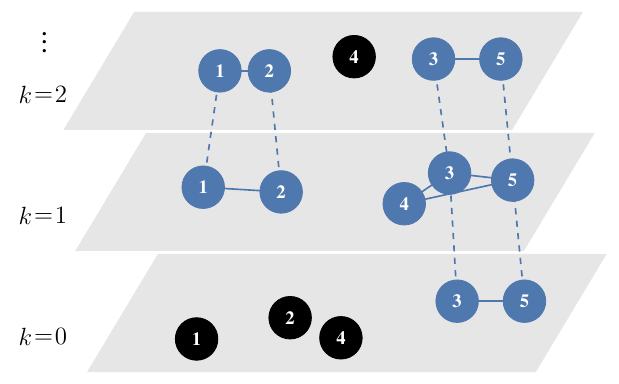}
    \centering
    \caption{Graphical illustration of the collision graph (blue). Spatial edges (solid) and temporal edges (dashed) form collision clusters (shaded). The black nodes correspond to robots that do not collide at a timestep $k$.}
    \label{fig:collision-graph}
\end{figure}
\subsection{Collision Graph}
\label{sec:collision-graph}
Given a trajectory $\tau_l = (p^{(i)}[k])^{i = 1,\ldots,N}_{k = 0, \ldots, K}$, we construct a spatio-temporal graph $\mathcal{G} = (\mathcal{V}, \mathcal{E}_s \cup \mathcal{E}_t)$; see \cref{fig:collision-graph}. A node $(k,i)\in\mathcal{V}$ is created for every robot $i$ at every timestep $k$. We connect vertices based on two conditions. If $\|{p}^{(i)}[k]-{p}^{(j)}[k]\|_2<R$, then $((k, i), (k, j)) \in \mathcal{E}_s$ (\emph{spatial edges}). If robot $i$ collides with at least one other robot at time $k$ and $k + 1$, then $((k, i), (k + 1, i)) \in \mathcal{E}_t$ (\emph{temporal edges}).
Spatial edges capture \emph{which} robots are involved in a collision and temporal edges capture \emph{for how long}. Each connected component of~$\mathcal{G}$ defines a \emph{collision cluster} $\mathcal{G}_{m} = (\mathcal{V}_m, \mathcal{E}_m), m = 1,\ldots, M$, with $M$ the number of identified clusters.
Building the graph requires detecting all pairwise violations across all time steps.  We use a per-timestep KD-tree data structure \cite{bentley1975multidimensional} to query all pairs within distance~$R$, avoiding the~$\mathcal{O}(N^2 K)$ cost of combinatorial checking.

\begin{figure}
    \centering
    \includegraphics[width=\linewidth]{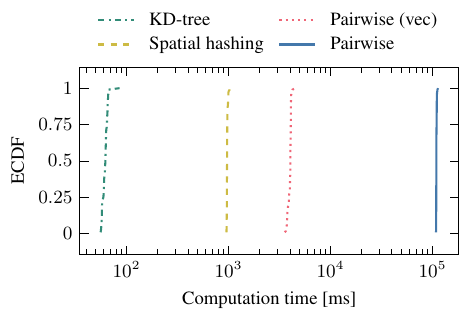}
    \vspace{-.8cm}
    \caption{Empirical cumulative distribution functions (ECDFs) of the time to compute the collision graph using different strategies. Here, $N = 500, K = 1000$; see \cref{exp:collision}.}
    \label{fig:compute-time-collision-graph}
\end{figure}

\begin{experiment}
\label{exp:collision}
    We build the collision graph for the linearly-interpolated initialization on $100$ randomized configurations with $N = 500$ vehicles, $K = 1000$ and $20\%$ allocated randomly (cf. \cref{sec:sim}).
    \emph{Results.} In \cref{fig:compute-time-collision-graph}, we report the cumulative time to compute the collision detection over all time steps using a KD-tree and compare it with the time it would take if we were naively checking all pair-wise collisions (with and without vectorization) or using a spatial hashing (i.e., discretizing the plane into uniform grid cells and hashing each robot into a cell so that collision checks are restricted to robots in the same or adjacent cells). This experiment clearly indicates the advantages of using a KD-tree with an average speedup of nearly $2000\times$ with respect to the pairwise collision checking.
\end{experiment}

% -----------------------------------------------------------------------
\subsection{Clustering}
\label{sec:clustering}
Collision clusters span only the timesteps at which violations occur. For effective re-planning, we expand each robot's temporal window within a cluster to give the optimizer additional freedom to plan collision-free detours. Concretely, for each robot $i$ in a collision cluster $\mathcal{G}_m = (\mathcal{V}_m, \mathcal{E}_m)$, let $k_{\min}^{(i)} = \min\{k \mid (k, i) \in \mathcal{V}_m\}$ and $k_{\max}^{(i)} = \max\{k \mid (k, i) \in \mathcal{V}_m\}$ be the earliest and latest collision timesteps of robot~$i$. We define the \emph{buffered timeframe} for robot~$i$ as $[\hat{k}_{\min}^{(i)}, \hat{k}_{\max}^{(i)}] = (\max(0, k_{\min}^{(i)} - B), \ldots,
    \min(K, k_{\max}^{(i)} + B))$,
where $B \in \mathbb{N}$ is a hyper-parameter; in our work, we fix it to the number of discrete time steps that correspond to $3$ seconds. Each robot in a cluster thus carries an individual time window, and the expanded cluster $\hat{\mathcal{G}}_m$ is defined over the union of these per-robot windows.

This buffering step can create overlaps between expanded clusters: there may exist $m \neq m'$ and a robot~$i$ such that $i$ appears in both $\hat{\mathcal{G}}_m$ and $\hat{\mathcal{G}}_{m'}$ with overlapping buffered timeframes, so that solving the two subproblems independently would produce conflicting trajectory updates. We detect all such overlaps and merge the corresponding clusters using a union-find procedure~\cite{cormen2022introduction}. However, naively merging all overlapping clusters can yield subproblems that are too large and negate the benefit of parallelization. We therefore estimate the resulting subproblem complexity as $S(\hat{\mathcal{G}}) = \mathrm{nnz}(P) + \mathrm{nnz}(C)$, the number of nonzero entries in the objective and constraint matrices of the linearized QP; see Section~\ref{sec:subproblem-solve}. If merging two clusters would produce a component with $S(\hat{\mathcal{G}}) > S_{\max}$, we split the group. Let $\hat{\mathcal{G}}_1, \ldots, \hat{\mathcal{G}}_P$ denote the buffered subproblems whose union exceeds the threshold. For each candidate $p \in \{1, \ldots, P\}$, we take $\hat{\mathcal{G}}_p$ out and proceed with the clustering of the others, obtaining a set of clusters $\hat{\mathcal{G}}_1', \ldots, \hat{\mathcal{G}}_{M(p)}'$. We score the candidate $p$ with $S_p^* = {\max} (S(\hat{\mathcal{G}}'_1), \ldots, S(\hat{\mathcal{G}}'_{M(p)}))$. We then keep $p^* = \arg\min_p S_p^*$ as a singleton and proceed with clustering. We repeat this procedure until all components satisfy $S \leq S_{\max}$ or are singletons. %Removed subproblems are deferred to the next outer iteration and will be identified automatically by the collision detection.

\begin{experiment}
\label{exp:clustering}
On the configurations of Experiment~\ref{exp:collision}, we run the full pipeline without buffering and without the cap on subproblem complexity, comparing success rate and computation time against the baseline SCP algorithm.
\emph{Results.} Without buffering, the pipeline does not converge. Without the cap, subproblems merge uncontrollably in congested situations, resulting in intractable subproblems. The pipeline's robustness and performance can drop up to 30\%.
\end{experiment}

% -----------------------------------------------------------------------
\subsection{Subproblem solving}
\label{sec:subproblem-solve}
Next, we look at how to solve the resulting subproblems, each described by a cluster $\hat{\mathcal{G}}_m$.
We adopt the \gls*{scp} framework of \citet{augugliaro_generation_2012} for generating collision-free trajectories. The trajectory of robot~$i$ is affine in its control inputs; velocities~${v}^{(i)}[k]$ for the single-integrator model. Collecting all control inputs into a single vector~${\chi} \in \mathbb{R}^{2NK}$, the trajectory optimization takes the form of a \gls*{qp}:
\begin{equation}
    \min_{{\chi}} \;\|{\chi}\|_2^2 \quad \text{s.t.} \quad {l} \leq C\,{\chi} \leq {u},
    \label{eq:qp}
\end{equation}
with the constraint matrix ~$C = [C_\mathrm{box}^\top \; C_\mathrm{coll}^\top]^\top$. The box inequality rows~$C_\mathrm{box}$ encode dynamics, initial and final position constraints, and box limits on the state and control inputs. The collision rows~$C_\mathrm{coll}$ enforce the non-convex collision avoidance constraint via a first-order Taylor expansion
\begin{equation}
\begin{aligned}
    &{\hat{n}}_{ij}[k]^\top \bigl({p}^{(i)}[k] - {p}^{(j)}[k]\bigr) \geq R, \\
    &{\hat{n}}_{ij}[k] = \frac{\bar{{p}}^{(i)}[k] - \bar{{p}}^{(j)}[k]}{\left\|\bar{{p}}^{(i)}[k] - \bar{{p}}^{(j)}[k]\right\|},
\end{aligned}
\label{eq:lin-collision}
\end{equation}
where~$\bar{{p}}$ denotes the previous iterate and linearization point, and $R$ is the collision distance. The \gls*{qp}~\eqref{eq:qp} is solved iteratively using OSQP \cite{stellato_osqp_2020} at default (tolerances: $10^{-4}$) accuracy  with updated linearizations~\eqref{eq:lin-collision} until the trajectory improvement satisfies the prescribed tolerances \cite{malyuta2022convex}; in this work we consider modest SCP accuracies (tolerances: $10^{-3}$).

In large-scale scenarios, certain subproblems can become numerically ill-conditioned, for instance in symmetric configurations. We mitigate this by perturbing the SCP initialization and, if a solve is infeasible, re-solving the subproblem with an adaptively coarsened discretization rate. The resulting trajectory is then interpolated to the global discretization and refined via a lightweight SCP polishing step, significantly increasing robustness for difficult configurations with labelled fractions above $10\%$.

\begin{figure}[t]
    \centering
    \includegraphics[width=\linewidth]{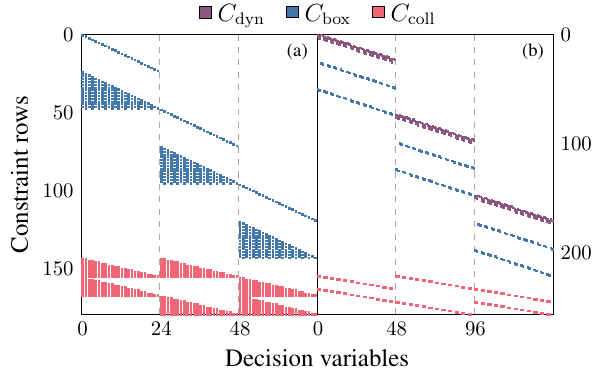}
    \vspace{-.75cm}
    \caption{Sparsity pattern of the single-integrator constraint matrix $C$ for $N=3$ robots and $K=12$ time steps. (a)~Direct shooting with $\chi = v$: $\mathrm{nnz} = 1476$. (b)~Direct transcription with $\chi = [p^\top,\, v^\top]^\top$: $\mathrm{nnz} = 432$. Dashed lines separate per-robot variable blocks.}
    \label{fig:sparsity}
\end{figure}

In the \emph{direct shooting} formulation employed by \citet{augugliaro_generation_2012} and described so far, positions are obtained by integrating the full control history,
$p^{(i)}[k] = p_0^{(i)} + h \sum_{m=0}^{k-1} v^{(i)}[m]$.
Thus, each collision constraint row at timestep~$k$ couples \emph{all}~$k$ preceding control inputs of both robots involved. Consequently, ~$C_\mathrm{coll}$ comprises lower-triangular blocks, with~$\mathcal{O}(N^2 K^2)$ non-zero entries.
However, it is possible to substantially reduce the number of non-zero entries (and, thus, increase the efficiency of the OSQP solver) by adopting a \emph{direct transcription} formulation 
that augments the decision vector:~${\chi} = [{p}^{(i)\top}\![k],\, 
{v}^{(i)\top}\![k]]_{i,k}^\top \in 
\mathbb{R}^{4NK}$. The dynamics become local equality constraints $C_\mathrm{dyn}$ between consecutive time steps, and each collision 
row~\eqref{eq:lin-collision} can be directly encoded in the pairwise positions. The dynamics constraints are then locally banded; see \cref{fig:sparsity}. The complexity reduces from~$O(N^2 K^2)$ to~$O(N^2 K)$ at the cost of a larger decision vector. 

\begin{figure}
    \centering
    \includegraphics[width=\linewidth]{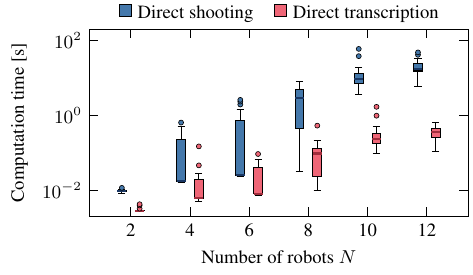}
    \vspace{-.75cm}
    \caption{Comparison of the time to solve a subproblem for varying number of robots between the direct shooting ($\chi = v$) and the direct transcription ($\chi = [p^\top,\, v^\top]^\top$) SCP formulations. Here, $K = 100$; see \cref{exp:scp}.}
    \label{fig:transcriptions}
\end{figure}

\begin{experiment}
\label{exp:scp}
We compare the \emph{direct shooting} and the \emph{direct transcription} formulations on randomized configurations to assess the performance. For $N\in \{2,4,6,8,10,12\}$ and a time horizon of $T=10\sqrt{N/2}$, we sample the initial and final positions for each robot uniformly inside a ball of radius $r(N)=5\sqrt{N/2}$. We repeat the experiment with $50$ different seeds. The time steps are kept constant $K=100$ and all robots are labelled. Cluster dimensions are scaled so that the congestion density is maintained across the experiments. We run the experiment for the modest SCP tolerance.
\emph{Results.} We report the statistics of the computation time for the two methods across varying numbers of robots in \cref{fig:transcriptions}. Overall, the direct transcription formulation improves over the direct shooting by up to two orders of magnitude for the larger problem instances and is the one that we adopt.
Crucially, the monolithic SCP becomes intractable for hundreds of robots, motivating the hierarchical decomposition.
\end{experiment}


\subsection{Merging}
\label{sec:merging}
After solving all subproblems in parallel, the local
solutions are merged back into the global trajectory by overwriting each
robot's state variables in its time window. The pipeline loop then rebuilds the collision graph and repeats. However, this procedure can introduce collisions elsewhere and possibly result in the pipeline oscillating on identical collision patterns. We detect such cycles by recording the signature $(\mathcal{R}_m, [k_s, k_e])$ of every subproblem at each iteration $\ell$, where $\mathcal{R}_m \subseteq \{1, \ldots, N\}$ and $\{k_s, \ldots, k_e\}$ is the time range of the subproblem. A subproblem at iteration $\ell$ is declared cyclic if there exists an earlier iteration $\ell' < \ell$ containing a subproblem with the same robot set $\mathcal{R}_m$ and an overlapping time range $\{k_s, \ldots, k_e\} \cap \{k'_s, \ldots, k'_e\} \neq \emptyset$.
Upon cycle detection, all subproblems with robot set $\mathcal{R}_j$ satisfying $\mathcal{R}_j \cap \mathcal{R}_m \neq \emptyset$ are locked into a single joint subproblem for all iterations $\ell''\geq \ell$.

\subsection{Comparison with SCP}
With hundreds of robots, the setting of interest of this paper, the baseline SCP algorithm is computationally too expensive, preventing a direct comparison with our complete pipeline. Thus, we defer a numerical case study with 500 robots to \cref{sec:sim}, and here compare our solver with the baseline SCP algorithm with 20 robots. In particular, we compare the two methods in scenarios of varying congestion, a property that our hierarchical approach exploits.

\begin{figure}
    \centering
    \includegraphics[width=\linewidth]{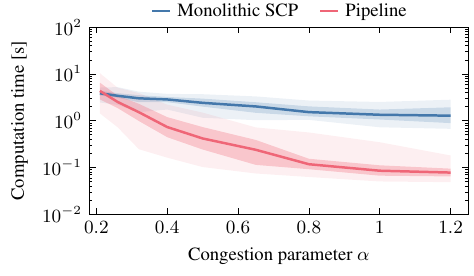}
    \vspace{-.75cm}
    \caption{Computation time of the hierarchical pipeline and the monolithic SCP baseline versus congestion density for $N = 20$ (tractable for the baseline SCP algorithm); see \cref{exp:congestion}. Lines mark median over $50$ seeds, with interquartile and $95\%$ bands.}
    \label{fig:congestion}
\end{figure}

\begin{experiment}
\label{exp:congestion}
    We assess the effect of the hierarchical decomposition by comparing the full pipeline against the monolithic SCP baseline. 
    We vary the congestion density $\alpha$: decreasing $\alpha$ linearly shrinks the workspace, making the configurations more congested. 
    \emph{Results.} Fig.~\ref{fig:congestion} reports the computation times of both solvers. Under heavy congestion the collision graph is dense, and the pipeline merges its subproblems until it solves essentially the monolithic problem; its runtime is then of the same order as the baseline, with an overhead from graph construction, clustering, and the outer loop. As congestion eases, the monolithic cost changes little while the pipeline cost falls by over an order of magnitude.
\end{experiment}

\section{Examples}
\label{sec:numerical-results}
In this section, we provide representative examples in simulation and real-world deployments to illustrate the qualitative and quantitative behavior of the proposed pipeline. All the data is collected on a MacBook Pro with an M3 Pro chip and 32GB of memory.

\subsection{Case study: Motion planning for 500 crafts.}
\label{sec:sim}
We consider a large-scale case study with $N=500$ vehicles, $K=1000$ timesteps and labelled robots from $0\%$ to $100\%$. The partial ``manual'' allocation is uniform at random over the robots and their target locations. As keyframe configurations we use a star, a heart, and the words ``Way'', ``Water'' and ``of''.
For each pattern, vehicle positions are sampled by placing $N$ points with uniform spacing of $4.5~\mathrm{m}$ along the corresponding curve; the perimeters of the shapes are thus calculated to ensure uniform spacing. As the shapes have varying sizes we scale time horizons accordingly; the discretization rate is evaluated based on $K$ and the maximum distance any robot has to travel.  
\begin{figure}
    \centering
    \includegraphics[width=\linewidth]{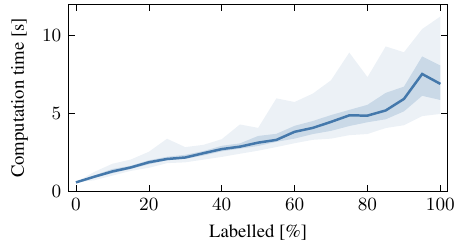}
    \vspace{-.75cm}
    \caption{Computation times for the 500-robot case study across ten keyframe transitions and varying fractions of labelled robots; see \cref{sec:sim}. The line marks medians over $50$ seeds, with interquartile and $95\%$ bands.}
    \label{fig:sim:times}
\end{figure}
The position box is sized with a $40\%$ margin around the larger of the two initial and final shapes. We report the statistics over 50 planning repetitions for each ratio of labelled robots and each of the initial-terminal configurations in \cref{fig:sim:times}. For each repetition we consider a different random seed to sample the ``manual'' allocation. 
The solver appears to be robust, with $100\%$ convergence rate up to $30\%$ labelled robots. For larger amounts of labelled robots the solver succeeds in $99\%$ of the instances within a $60\,\mathrm{s}$ timeout. 
Overall, the speed of our planner, which is able to successfully find collision-free trajectories within seconds, allows for a smooth user experience during choreography planning. 
We visualize representative trajectories in \cref{fig:sim}, showing time snapshots of the fleet as well as the aggregate spatial footprint of the motion over the full horizon. In general, the trajectories span a few hundred meters and take several minutes to complete.

\subsection{Case study: Deployment of 24 robots on Lake Zürich.}
\label{sec:zurich}
For the demonstration on Lake Z\"urich, we deployed a fleet of $24$ robots and choreographed transitions for $16$ robots spanning one to two minutes. Using our planner within the choreography tool, designers could iteratively adjust keyframes and re-synthesize collision-free, dynamically feasible trajectories within seconds per design iteration, despite individual transitions discretizing into more than 400 time steps. A live demonstration picture is shown in \cref{fig:planning-cover}, and one of the performed trajectories in Example 3 in \cref{fig:sim}.

\subsection{Case study: Deployment at the Time Space Existence 2025 Venice Biennale.}
\begin{figure}
    \centering
    \includegraphics[width=\linewidth]{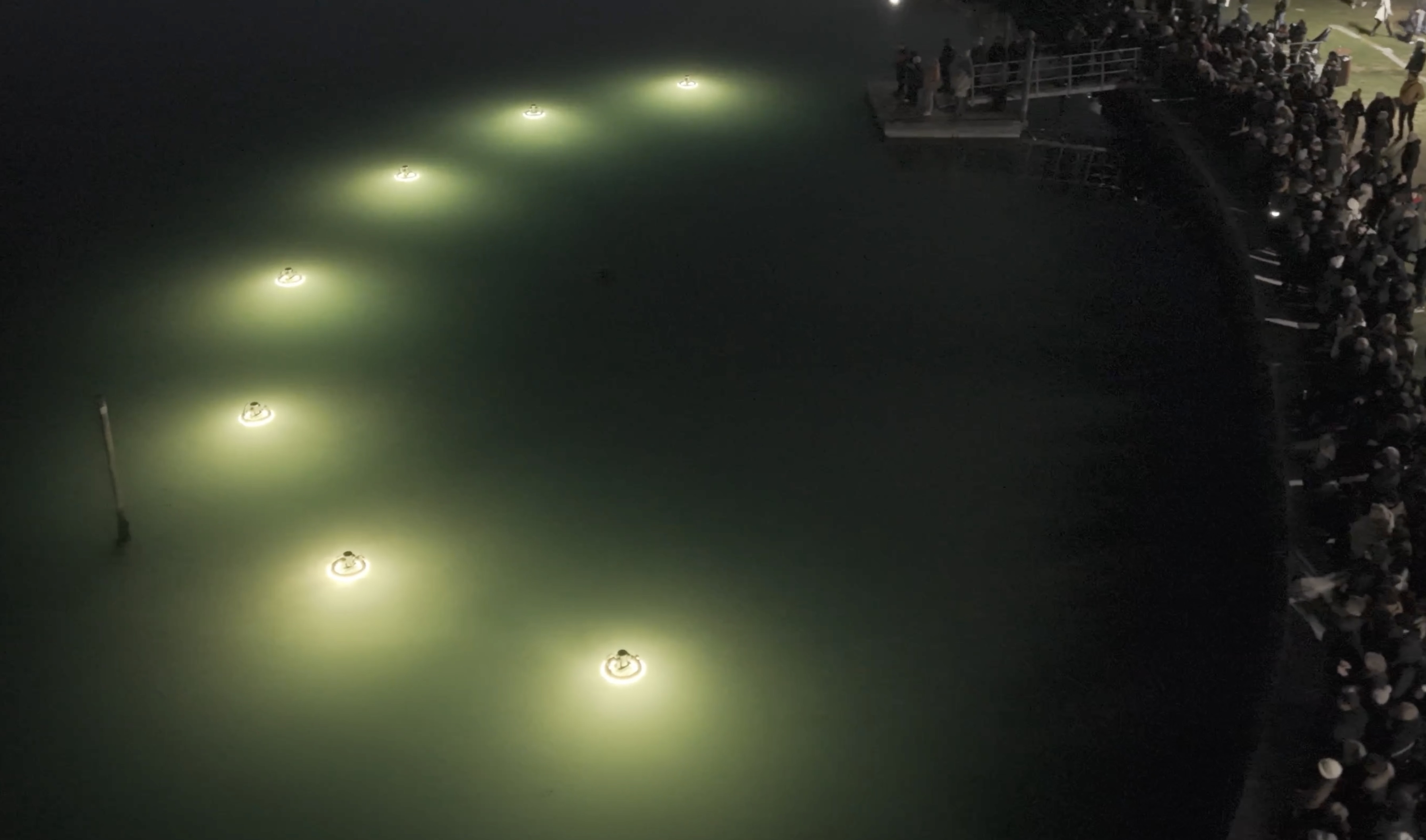}
    \vspace{-.5cm}
    \caption{Floating robots in formation at the Time Space Existence 2025 Venice Biennale.}
    \label{fig:venice}
\end{figure}

\begin{figure*}
    \centering
    \begin{tcolorbox}[
      colback=white,
      colframe=blue!20,
      arc=1mm,
      boxsep=0pt,
      top=2pt,
      left=5pt,
      right=5pt,
      bottom=2pt,
      toptitle=3pt,
      bottomtitle=2pt,
      coltitle=black,
      title=\centering\textbf{Example 1: $N = 500, K = 1000$, from "WATER" to "WAY"}
    ]
    \begin{minipage}{\textwidth}
        \includegraphics[width=0.24\linewidth]{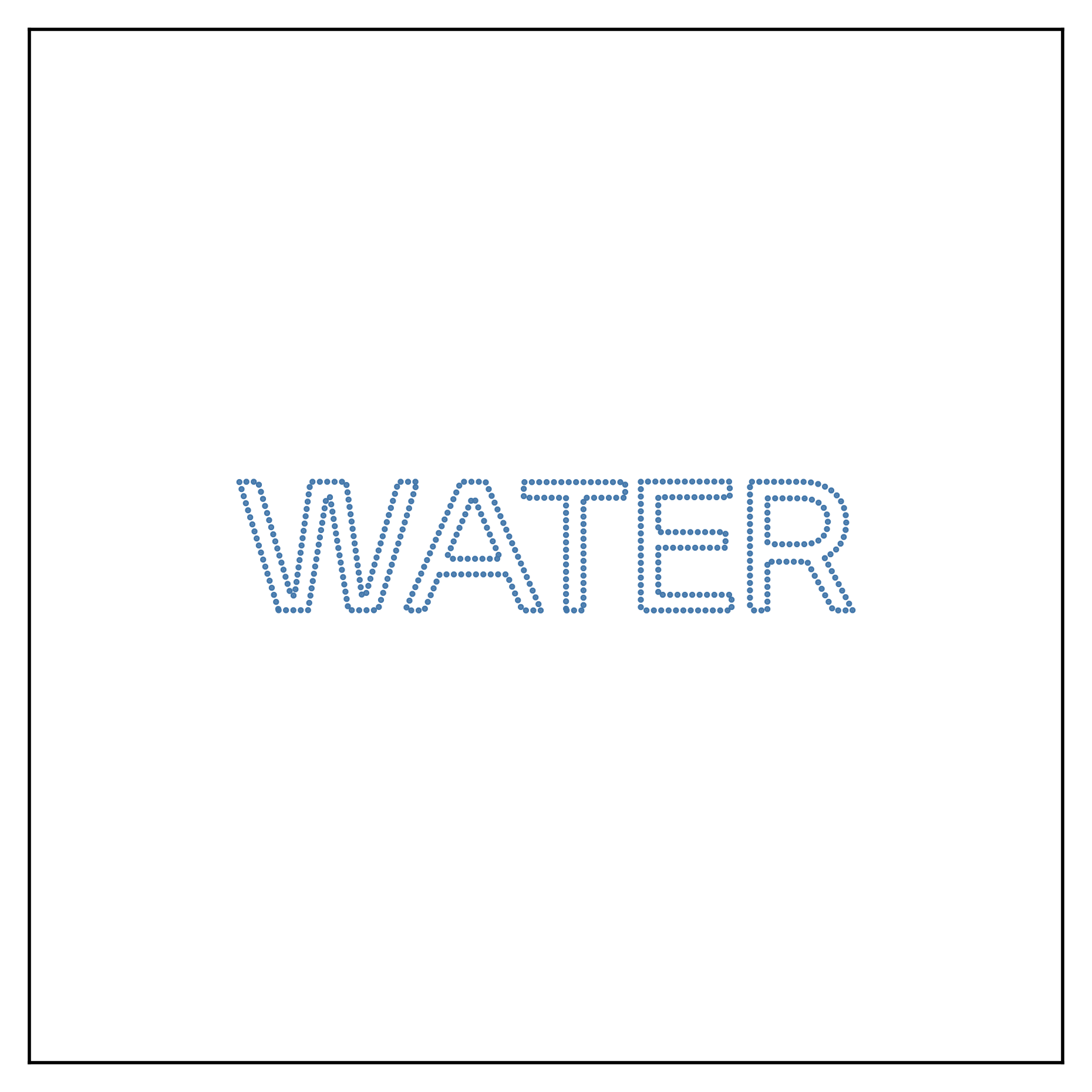}
        \hfill
        \includegraphics[width=0.24\linewidth]{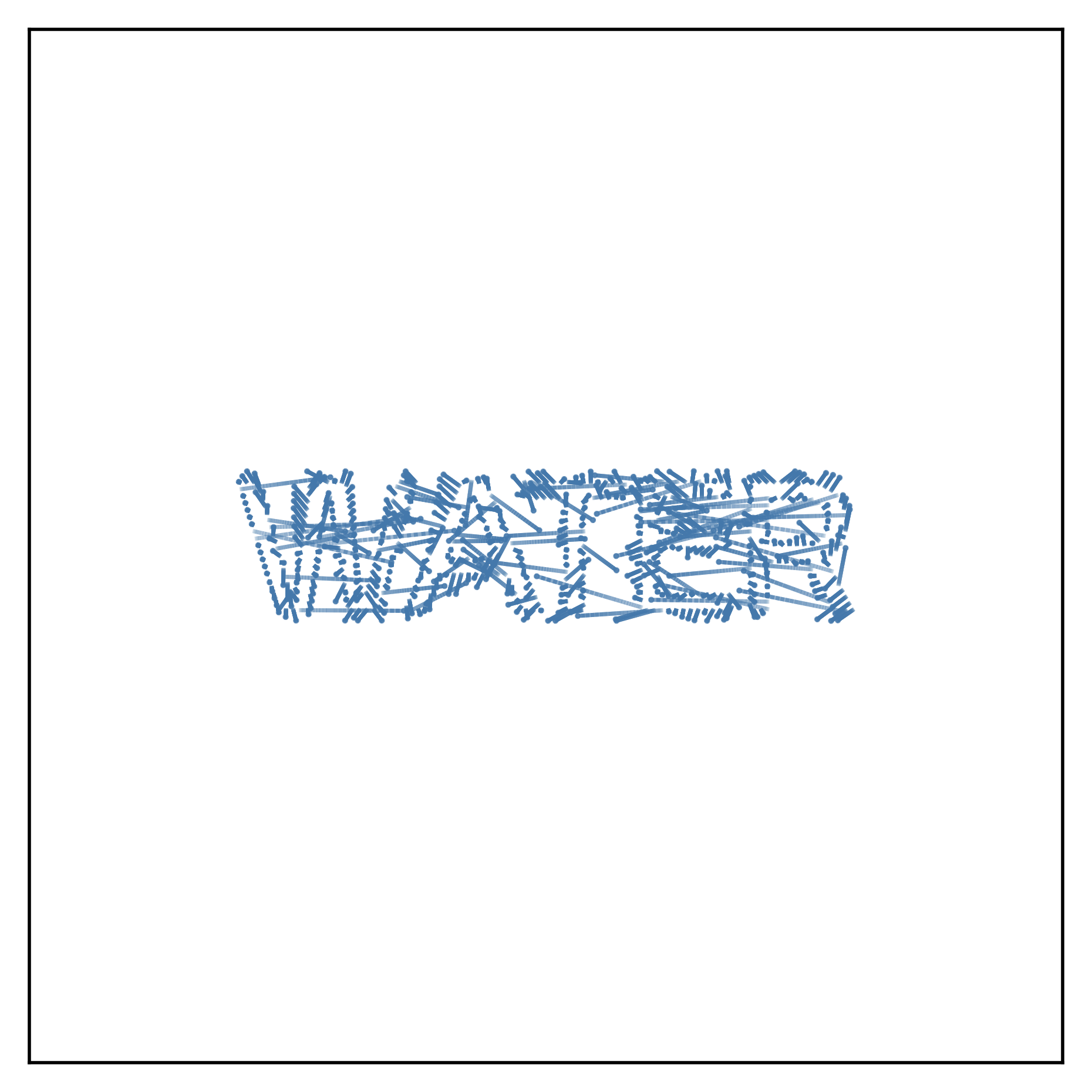}
        \hfill
        \includegraphics[width=0.24\linewidth]{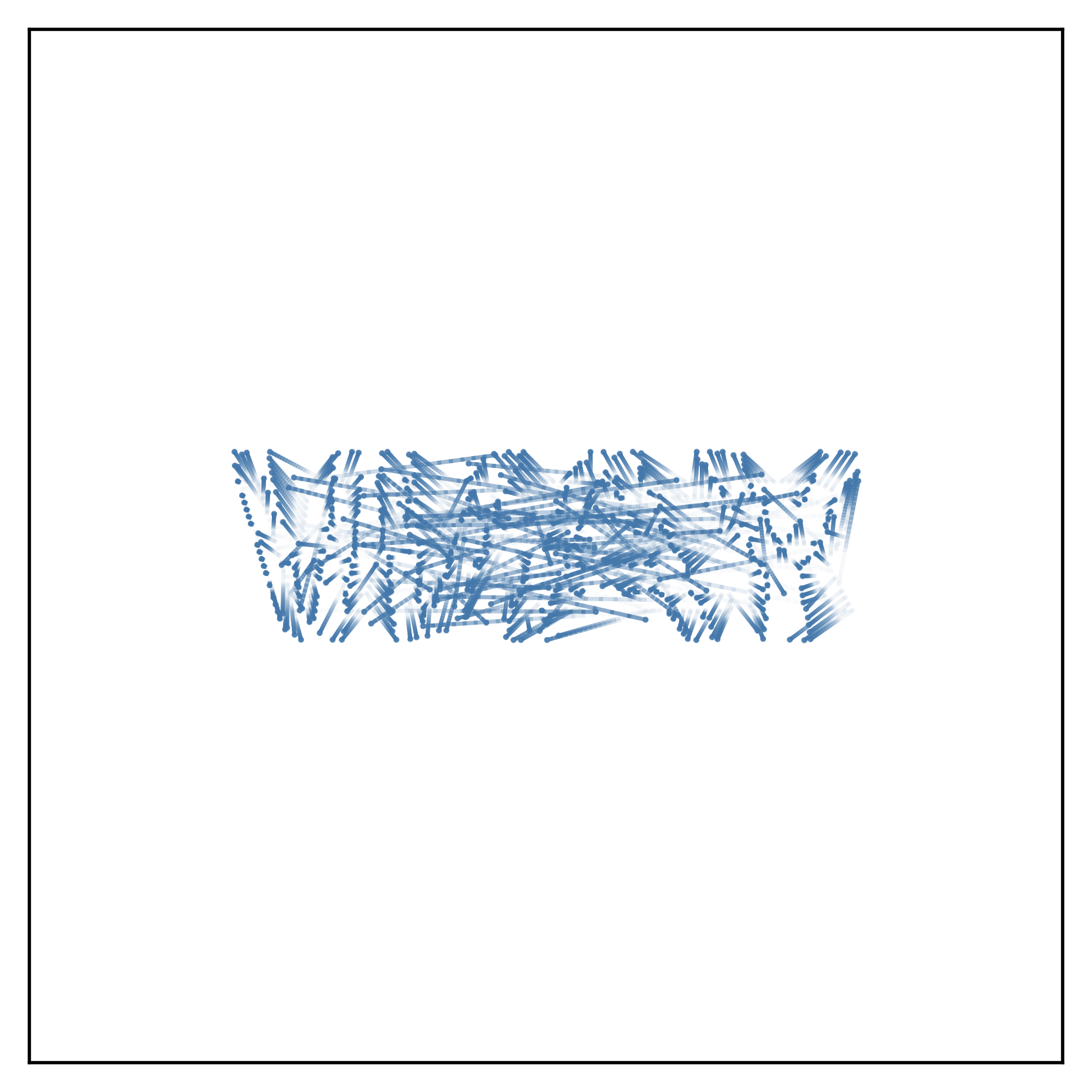}
        \hfill
        \includegraphics[width=0.24\linewidth]{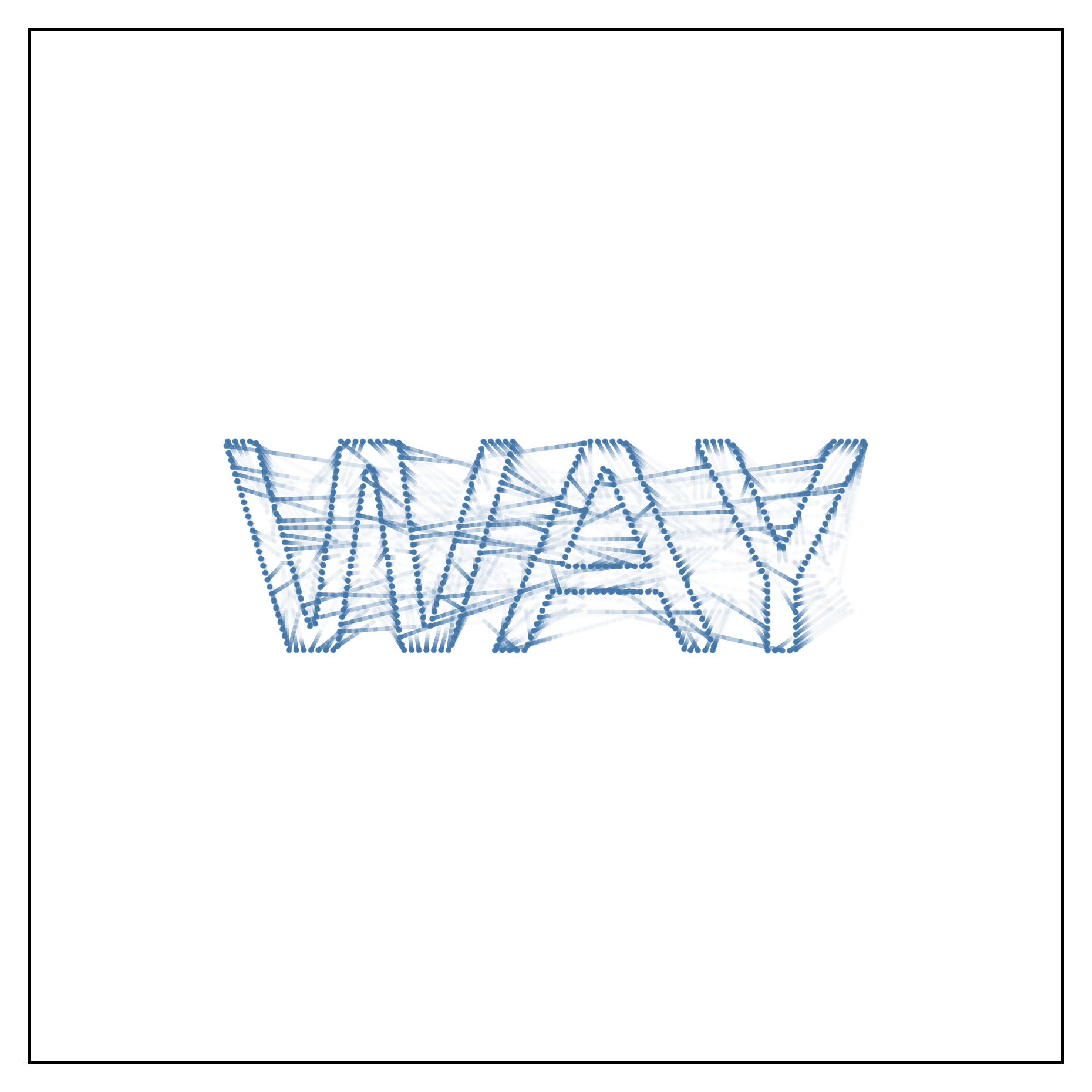}
    \end{minipage}
    \end{tcolorbox}
    \begin{tcolorbox}[
      colback=white,
      colframe=blue!20,
      arc=1mm,
      boxsep=0pt,
      top=2pt,
      left=5pt,
      right=5pt,
      bottom=2pt,
      toptitle=3pt,
      bottomtitle=2pt,
      coltitle=black,
      title=\centering\textbf{Example 2: $N = 500, K = 1000$, from "of" to star}
    ]
    \begin{minipage}{\textwidth}
        \includegraphics[width=0.24\linewidth]{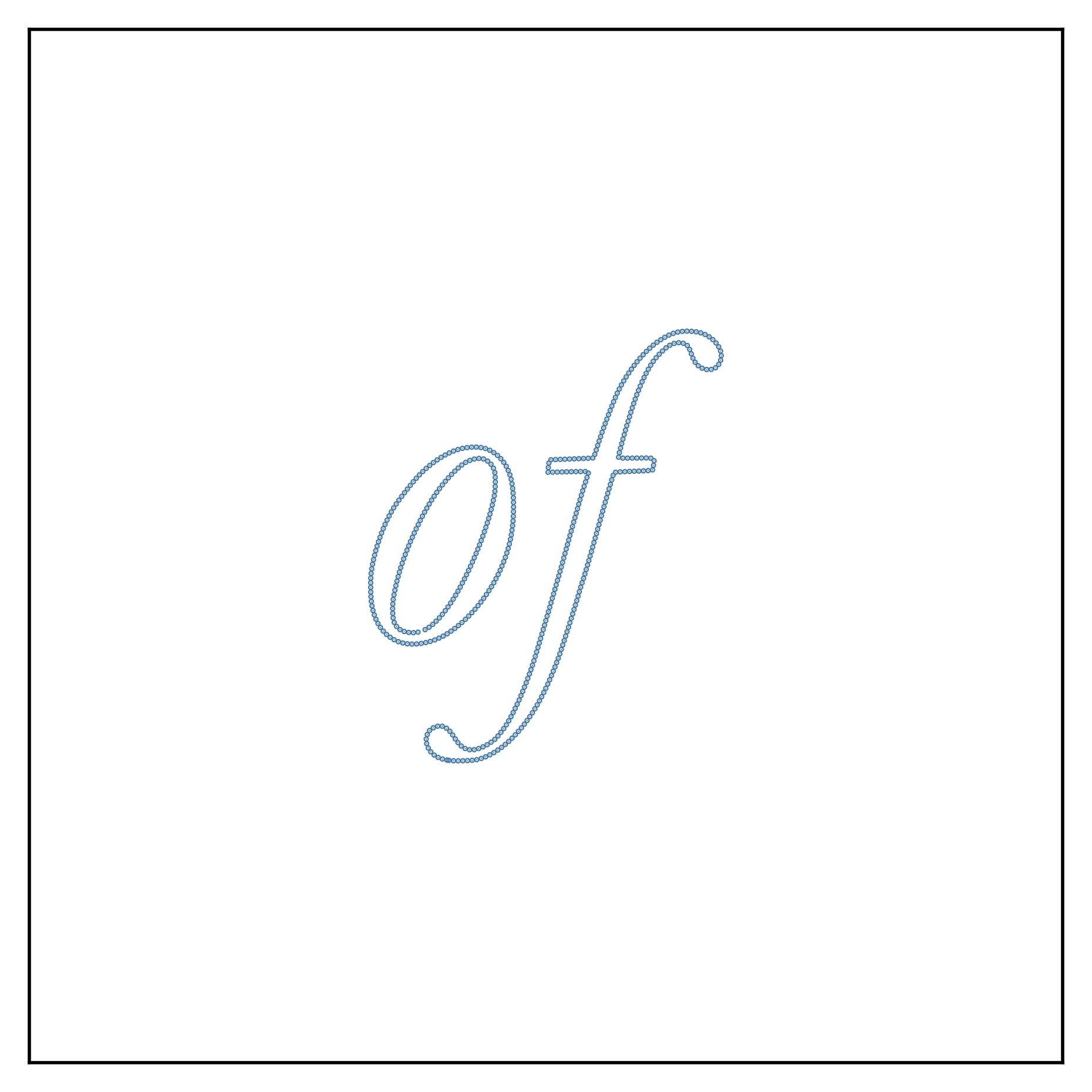}
        \hfill
        \includegraphics[width=0.24\linewidth]{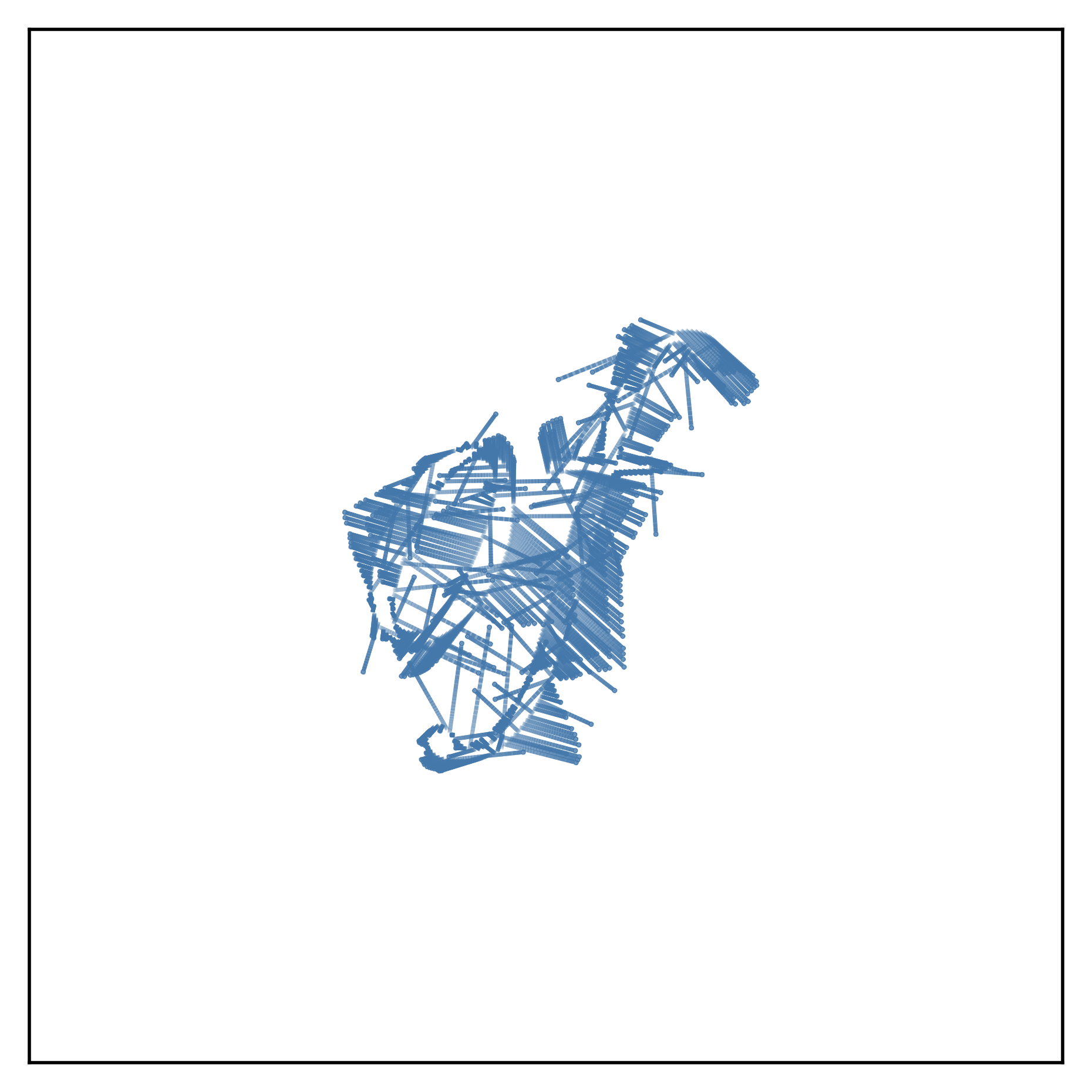}
        \hfill
        \includegraphics[width=0.24\linewidth]{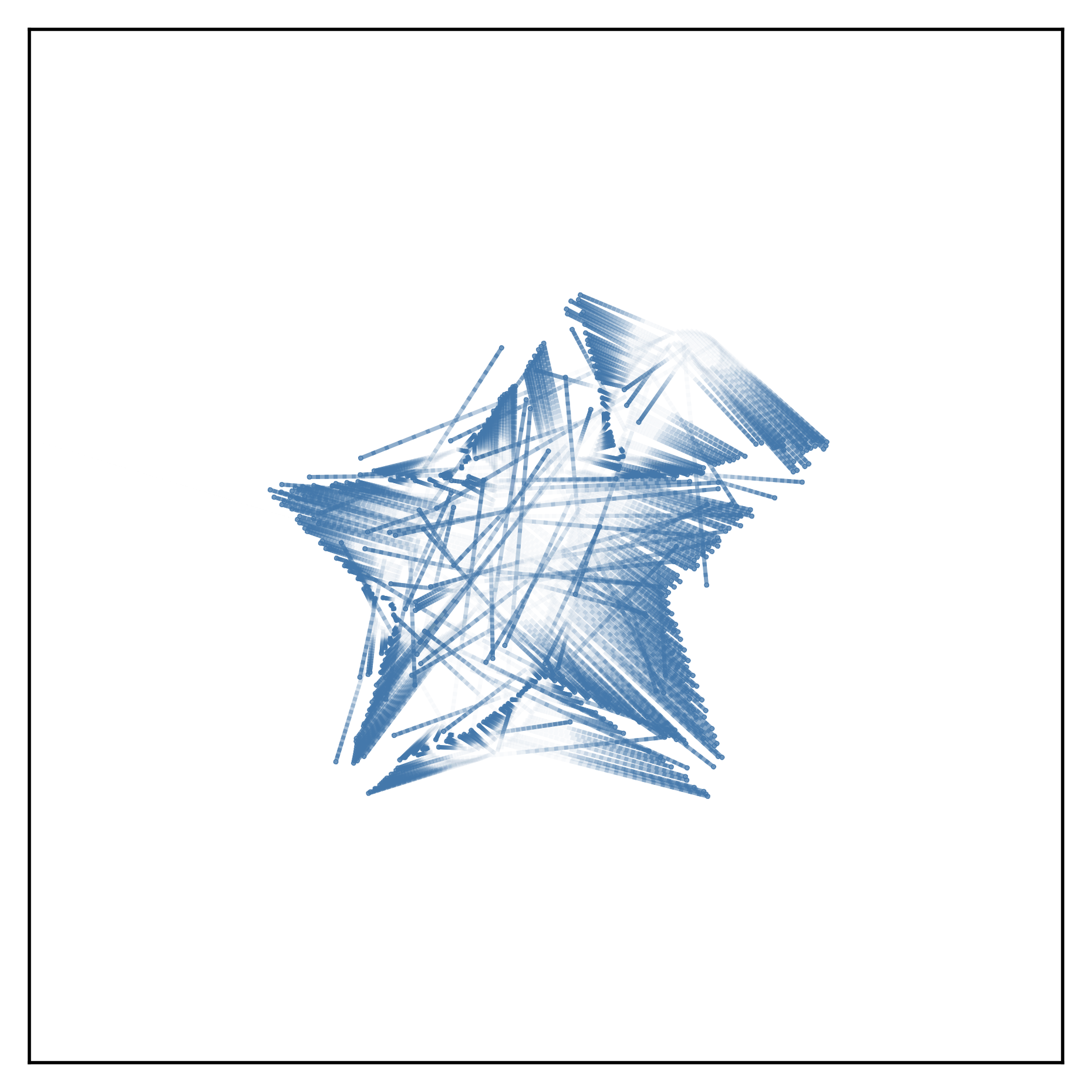}
        \hfill
        \includegraphics[width=0.24\linewidth]{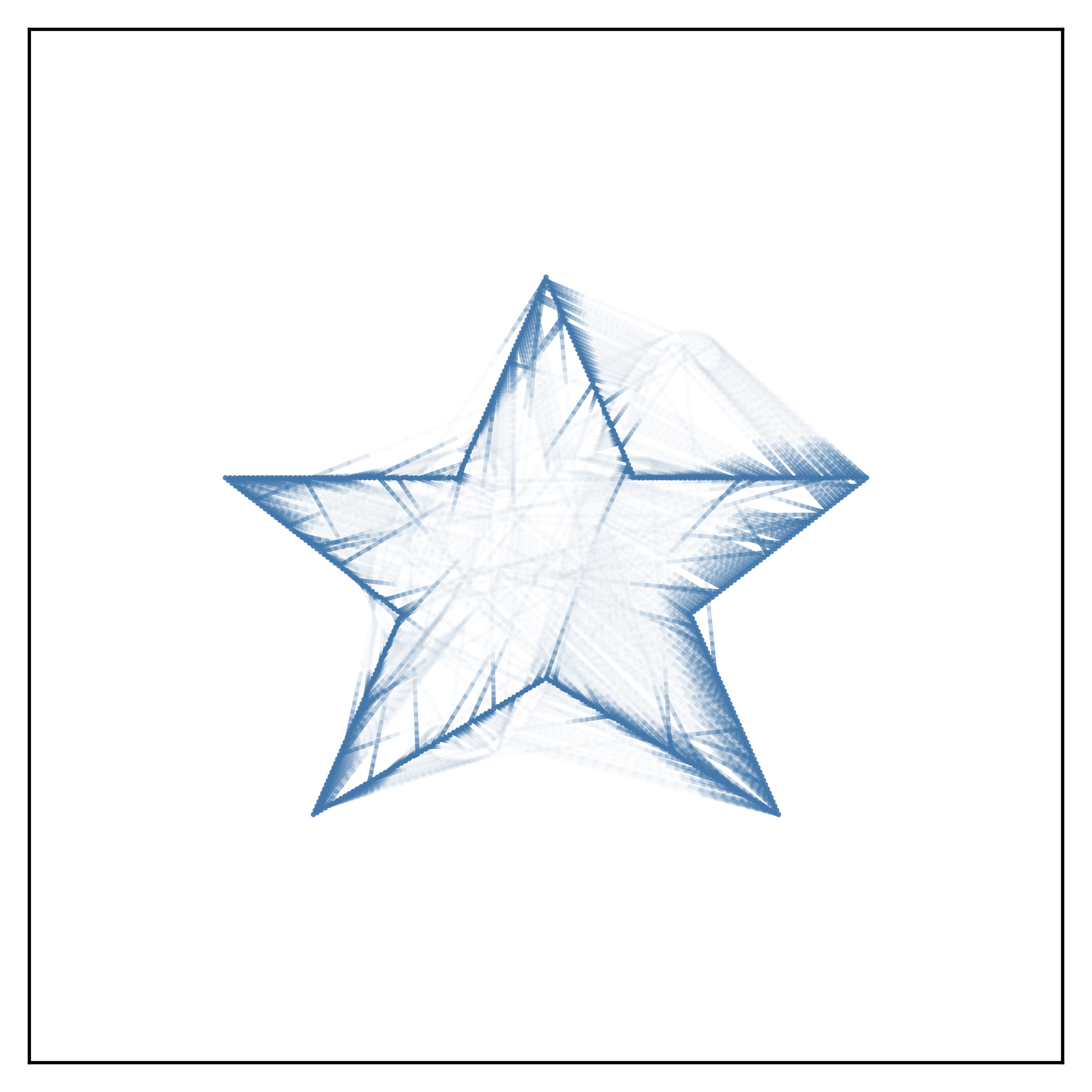}
    \end{minipage}
    \end{tcolorbox}
    \begin{tcolorbox}[
      colback=white,
      colframe=blue!20,
      arc=1mm,
      boxsep=0pt,
      top=2pt,
      left=5pt,
      right=5pt,
      bottom=2pt,
      toptitle=3pt,
      bottomtitle=2pt,
      coltitle=black,
      title=\centering\textbf{Example 3: The Lake Z\"urich demonstration}
    ]
    \begin{minipage}{\textwidth}
        \includegraphics[width=0.24\linewidth]{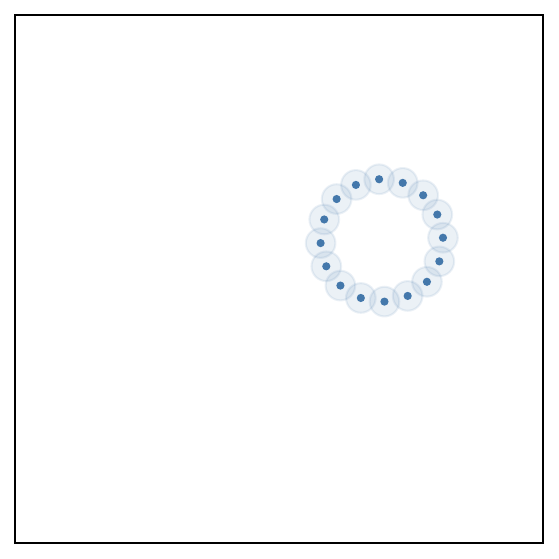}
        \hfill
        \includegraphics[width=0.24\linewidth]{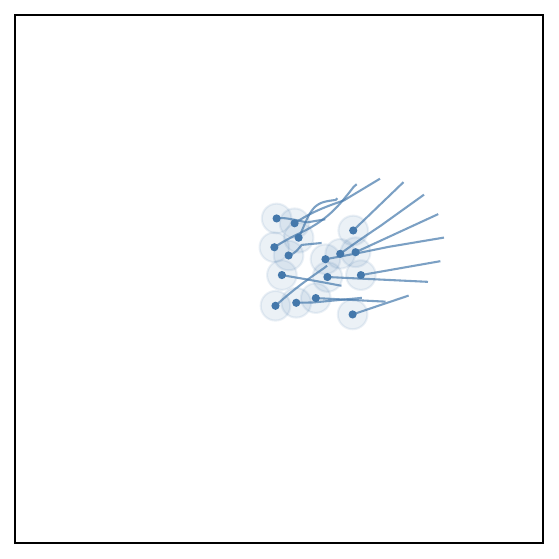}
        \hfill
        \includegraphics[width=0.24\linewidth]{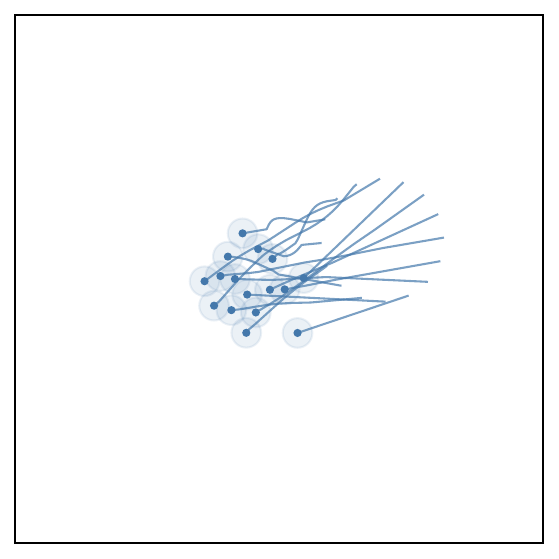}
        \hfill
        \includegraphics[width=0.24\linewidth]{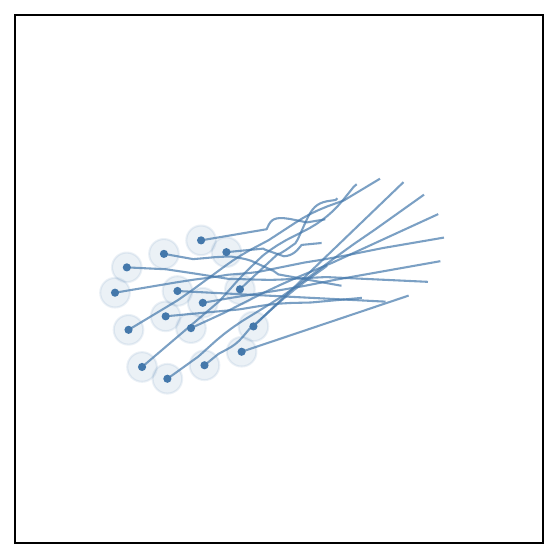}
    \end{minipage}
    \end{tcolorbox}
    \begin{tcolorbox}[
      colback=white,
      colframe=blue!20,
      arc=1mm,
      boxsep=0pt,
      top=2pt,
      left=5pt,
      right=5pt,
      bottom=2pt,
      toptitle=3pt,
      bottomtitle=2pt,
      coltitle=black,
      title=\centering\textbf{Example 4: The Time Space Existence 2025 Venice Biennale demonstration}
      % title=\centering\textbf{Example 4: The demonstration in Venice}
    ]
    \begin{minipage}{\textwidth}
        \includegraphics[width=0.24\linewidth]{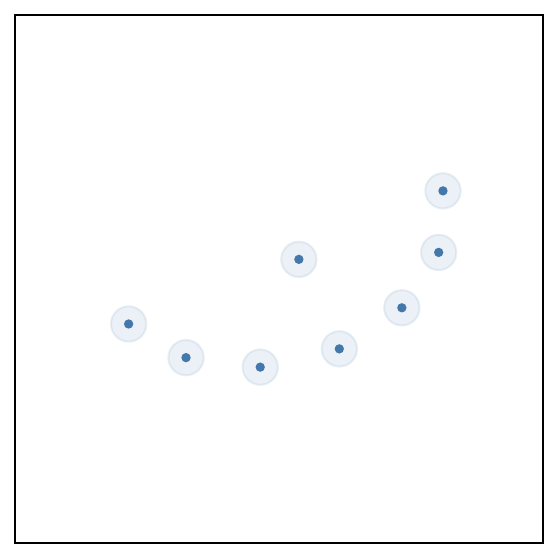}
        \hfill
        \includegraphics[width=0.24\linewidth]{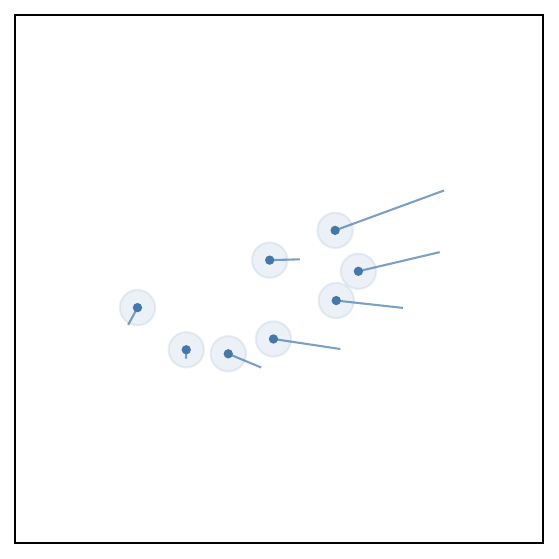}
        \hfill
        \includegraphics[width=0.24\linewidth]{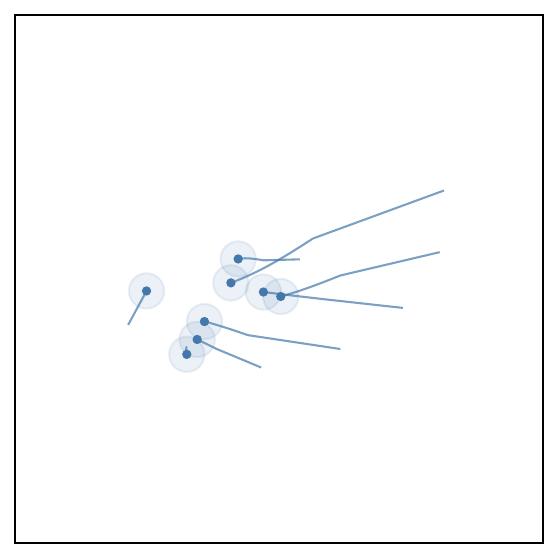}
        \hfill
        \includegraphics[width=0.24\linewidth]{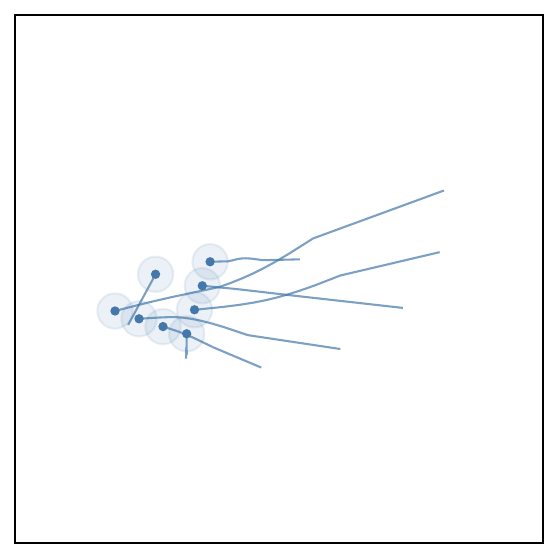}
    \end{minipage}
    \end{tcolorbox}
    \caption{Examples of synthesized trajectories. For each example, from left to right, each figure depicts a snapshot of the fleet at a different time-step, with a trailing highlighting the path covered so far. The right-most picture depicts the entire fleet trajectory over all timesteps.}
    \label{fig:sim}
\end{figure*}

% \subsection{Case study: Deployment in Venice.}
\label{sec:venice}
For the demonstration at the Time Space Existence 2025 Venice Biennale, we deployed a fleet of $8$ robots and designed several transitions. Using our planner within the choreography tool, designers could iteratively adjust keyframes and regenerate collision-free, dynamically feasible trajectories within $1$ to $2$ seconds per design iteration, despite the transition discretizing into more than $2000$ time steps. The final, executed trajectory comprised $2491$ steps and was computed in $1.23~\mathrm{s}$ with the proposed planner. A picture from the live demonstration is shown in \cref{fig:venice}, and one of the performed trajectories in Example 4 in \cref{fig:sim}. To validate trajectory tracking, we simulated the Venice deployment in a ROS2 environment with disturbances, obtaining a tracking RMSE of 0.063 m, consistent with errors observed during the live deployment under wind and waves.

\section{Conclusion} 
\label{sec:conclusion}
\emph{Contributions.}
We studied the motion planning problem for large fleets of omnidirectional surface robots, motivated by interactive show design where a user iteratively edits sparse keyframes and expects collision-free trajectories within seconds, even when transitions span minutes and thousands of discretization steps. To address the core scalability bottleneck induced by collision constraints, we proposed an engineering-focused pipeline that (i) builds a collision graph from a fast yet effective initialization, (ii) decomposes the globally coupled problem into interaction clusters, and (iii) solves these clusters independently (and in parallel), while incorporating robustness mechanisms. At the cluster level, we further improved solve efficiency and robustness through a reformulation that increases the sparsity of the linearized \gls*{qp}s and a dynamic time-scale selection strategy.

We validated the pipeline in simulations up to 500 robots and 1000 time steps, and the synthesized trajectories have been deployed in two real-world demonstrations, on Lake Z\"urich and at the Time Space Existence 2025 Venice Biennale.
We engineered the pipeline with the support of extensive ablations that isolate the effect of each design choice. The resulting planner substantially reduces computation time compared to traditional monolithic \gls*{scp} methods and enables the fast turnaround needed for iterative choreography design.

\emph{Outlook.} 
On the optimization side, our cluster-level solver can be strengthened with alternative subproblem parameterizations and solution spaces---for instance, planning directly in a contact space, or leveraging learned priors while enforcing feasibility via \emph{hard}-constrained neural architectures \cite{grontas2026pinet} that keep safety and boundary conditions guaranteed. Both collision detection and subproblem solves can also be parallelized over GPUs. On the choreography side, the allocation step is a powerful handle for artistic intent: replacing the cost structure $c_{ij}$ with richer \cite{lanzetti2024variational,terpin_dynamic_2024}, designer-controllable (and potentially learning-based) costs could encode style and incorporate human preference feedback \cite{christiano2017deep}, matching the subjective notion of ``visually-appealing'' motion.

\bibliographystyle{IEEEtran}
\bibliography{references}

\begin{thebibliography}{10}
\providecommand{\url}[1]{#1}
\csname url@samestyle\endcsname
\providecommand{\newblock}{\relax}
\providecommand{\bibinfo}[2]{#2}
\providecommand{\BIBentrySTDinterwordspacing}{\spaceskip=0pt\relax}
\providecommand{\BIBentryALTinterwordstretchfactor}{4}
\providecommand{\BIBentryALTinterwordspacing}{\spaceskip=\fontdimen2\font plus
\BIBentryALTinterwordstretchfactor\fontdimen3\font minus \fontdimen4\font\relax}
\providecommand{\BIBforeignlanguage}[2]{{%
\expandafter\ifx\csname l@#1\endcsname\relax
\typeout{** WARNING: IEEEtran.bst: No hyphenation pattern has been}%
\typeout{** loaded for the language `#1'. Using the pattern for}%
\typeout{** the default language instead.}%
\else
\language=\csname l@#1\endcsname
\fi
#2}}
\providecommand{\BIBdecl}{\relax}
\BIBdecl

\bibitem{ramachandranvenkatapathy2026wayofwater}
A.~K. Ramachandran~Venkatapathy, C.~Golling, S.~Burmester, N.~Sendlhofer, R.~Jiang, J.~Kamm, and R.~D'Andrea, ``Choreographing the {Way of Water}: A computational framework for aquatic robotic art,'' in \emph{Proceedings of the International Conference on New Interfaces for Musical Expression (NIME '26)}, 2026.

\bibitem{alonso-mora_multi-robot_2015}
J.~Alonso-Mora, S.~Baker, and D.~Rus, ``Multi-robot navigation in formation via sequential convex programming,'' in \emph{2015 {IEEE}/{RSJ} {International} {Conference} on {Intelligent} {Robots} and {Systems} ({IROS})}.\hskip 1em plus 0.5em minus 0.4em\relax Hamburg, Germany: IEEE, 2015, pp. 4634--4641.

\bibitem{10.1145/2858036.2858353}
C.~Gebhardt, B.~Hepp, T.~N\"{a}geli, S.~Stev\v{s}i\'{c}, and O.~Hilliges, ``Airways: Optimization-based planning of quadrotor trajectories according to high-level user goals,'' in \emph{Proceedings of the 2016 CHI Conference on Human Factors in Computing Systems}.\hskip 1em plus 0.5em minus 0.4em\relax New York, NY, USA: Association for Computing Machinery, 2016, p. 2508–2519.

\bibitem{marks_technical_1978}
B.~R. Marks and G.~P. Wright, ``Technical {Note}—{A} {General} {Inner} {Approximation} {Algorithm} for {Nonconvex} {Mathematical} {Programs},'' \emph{Operations Research}, vol.~26, no.~4, pp. 681--683, 1978.

\bibitem{tran_application_2009}
Q.~{Tran Dinh} and M.~Diehl, ``An application of sequential convex programming to time optimal trajectory planning for a car motion,'' in \emph{Proceedings of the 48th IEEE Conference on Decision and Control (CDC) held jointly with the 2009 28th Chinese Control Conference}.\hskip 1em plus 0.5em minus 0.4em\relax Shanghai, China: IEEE, 2009, pp. 4366--4371.

\bibitem{augugliaro_generation_2012}
F.~Augugliaro, A.~P. Schoellig, and R.~D'Andrea, ``Generation of collision-free trajectories for a quadrocopter fleet: {A} sequential convex programming approach,'' in \emph{2012 {IEEE}/{RSJ} {International} {Conference} on {Intelligent} {Robots} and {Systems}}.\hskip 1em plus 0.5em minus 0.4em\relax Vilamoura-Algarve, Portugal: IEEE, 2012, pp. 1917--1922.

\bibitem{schulman_finding_2013}
J.~Schulman, J.~Ho, A.~Lee, I.~Awwal, H.~Bradlow, and P.~Abbeel, ``Finding {Locally} {Optimal}, {Collision}-{Free} {Trajectories} with {Sequential} {Convex} {Optimization},'' in \emph{Robotics: {Science} and {Systems} {IX}}.\hskip 1em plus 0.5em minus 0.4em\relax Robotics: Science and Systems Foundation, 2013.

\bibitem{schulman_motion_2014}
J.~Schulman, Y.~Duan, J.~Ho, A.~Lee, I.~Awwal, H.~Bradlow, J.~Pan, S.~Patil, K.~Goldberg, and P.~Abbeel, ``Motion planning with sequential convex optimization and convex collision checking,'' \emph{The International Journal of Robotics Research}, vol.~33, no.~9, pp. 1251--1270, 2014.

\bibitem{siciliano_reciprocal_2011}
J.~Van Den~Berg, S.~J. Guy, M.~Lin, and D.~Manocha, ``Reciprocal n-body collision avoidance,'' in \emph{Robotics Research}, ser. Springer Tracts in Advanced Robotics, C.~Pradalier, R.~Siegwart, and G.~Hirzinger, Eds.\hskip 1em plus 0.5em minus 0.4em\relax Springer, 2011, vol.~70, pp. 3--19.

\bibitem{sharon_conflict-based_2015}
G.~Sharon, R.~Stern, A.~Felner, and N.~R. Sturtevant, ``Conflict-based search for optimal multi-agent pathfinding,'' \emph{Artificial Intelligence}, vol. 219, pp. 40--66, 2015.

\bibitem{ratliff_chomp_2009}
N.~Ratliff, M.~Zucker, J.~A. Bagnell, and S.~Srinivasa, ``{CHOMP}: {Gradient} optimization techniques for efficient motion planning,'' in \emph{2009 {IEEE} {International} {Conference} on {Robotics} and {Automation}}.\hskip 1em plus 0.5em minus 0.4em\relax Kobe: IEEE, 2009, pp. 489--494.

\bibitem{kalakrishnan_stomp_2011}
M.~Kalakrishnan, S.~Chitta, E.~Theodorou, P.~Pastor, and S.~Schaal, ``{STOMP}: {Stochastic} trajectory optimization for motion planning,'' in \emph{2011 {IEEE} {International} {Conference} on {Robotics} and {Automation}}.\hskip 1em plus 0.5em minus 0.4em\relax Shanghai, China: IEEE, 2011, pp. 4569--4574.

\bibitem{terpin2022distributed}
A.~Terpin, S.~Fricker, M.~Perez, M.~H. de~Badyn, and F.~D{\"o}rfler, ``Distributed feedback optimisation for robotic coordination,'' in \emph{2022 American Control Conference (ACC)}.\hskip 1em plus 0.5em minus 0.4em\relax IEEE, 2022, pp. 3710--3715.

\bibitem{chen_decentralized_2017}
Y.~F. Chen, M.~Liu, M.~Everett, and J.~P. How, ``Decentralized non-communicating multiagent collision avoidance with deep reinforcement learning,'' in \emph{2017 {IEEE} {International} {Conference} on {Robotics} and {Automation} ({ICRA})}.\hskip 1em plus 0.5em minus 0.4em\relax Singapore: IEEE, 2017, pp. 285--292.

\bibitem{zhang_neural_2023}
S.~Zhang, K.~Garg, and C.~Fan, ``Neural graph control barrier functions guided distributed collision-avoidance multi-agent control,'' in \emph{Proceedings of The 7th Conference on Robot Learning}, ser. Proceedings of Machine Learning Research, vol. 229, 2023, pp. 2373--2392.

\bibitem{stellato_osqp_2020}
B.~Stellato, G.~Banjac, P.~Goulart, A.~Bemporad, and S.~Boyd, ``{OSQP}: an operator splitting solver for quadratic programs,'' \emph{Mathematical Programming Computation}, vol.~12, no.~4, pp. 637--672, Dec. 2020.

\bibitem{mirsky1963results}
L.~Mirsky, ``Results and problems in the theory of doubly-stochastic matrices,'' \emph{Zeitschrift f{\"u}r Wahrscheinlichkeitstheorie und verwandte Gebiete}, vol.~1, no.~4, pp. 319--334, 1963.

\bibitem{terpin_dynamic_2024}
A.~Terpin, N.~Lanzetti, and F.~Dörfler, ``Dynamic {Programming} in {Probability} {Spaces} via {Optimal} {Transport},'' \emph{SIAM Journal on Control and Optimization}, vol.~62, no.~2, pp. 1183--1206, 2024.

\bibitem{kuhn1955hungarian}
H.~W. Kuhn, ``The hungarian method for the assignment problem,'' \emph{Naval research logistics quarterly}, vol.~2, no. 1-2, pp. 83--97, 1955.

\bibitem{bentley1975multidimensional}
J.~L. Bentley, ``Multidimensional binary search trees used for associative searching,'' \emph{Communications of the ACM}, vol.~18, no.~9, pp. 509--517, 1975.

\bibitem{cormen2022introduction}
T.~H. Cormen, C.~E. Leiserson, R.~L. Rivest, and C.~Stein, \emph{Introduction to algorithms}.\hskip 1em plus 0.5em minus 0.4em\relax MIT press, 2022.

\bibitem{malyuta2022convex}
D.~Malyuta, T.~P. Reynolds, M.~Szmuk, T.~Lew, R.~Bonalli, M.~Pavone, and B.~A{\c{c}}{\i}kme{\c{s}}e, ``Convex optimization for trajectory generation: A tutorial on generating dynamically feasible trajectories reliably and efficiently,'' \emph{IEEE Control Systems Magazine}, vol.~42, no.~5, pp. 40--113, 2022.

\bibitem{grontas2026pinet}
P.~D. Grontas, A.~Terpin, E.~C. Balta, R.~D'Andrea, and J.~Lygeros, ``Pinet: Optimizing hard-constrained neural networks with orthogonal projection layers,'' in \emph{The Fourteenth International Conference on Learning Representations}, 2026.

\bibitem{lanzetti2024variational}
N.~Lanzetti, A.~Terpin, and F.~D{\"o}rfler, ``Variational analysis in the wasserstein space,'' \emph{arXiv preprint arXiv:2406.10676}, 2024.

\bibitem{christiano2017deep}
P.~F. Christiano, J.~Leike, T.~Brown, M.~Martic, S.~Legg, and D.~Amodei, ``Deep reinforcement learning from human preferences,'' \emph{Advances in neural information processing systems}, vol.~30, 2017.

\end{thebibliography}

\end{document}